\documentclass{article}
\usepackage{hyperref}
\usepackage{subfig}
\usepackage{spconf,amsmath,graphicx}
\usepackage[dvipsnames]{xcolor}
\def\textbf#1{{\bfseries #1}}

\usepackage{enumitem}
\setlist{nosep, leftmargin=14pt}

\usepackage{mwe} 
\usepackage{float}
\DeclareMathOperator{\sign}{sign}

\title{Scribble-based fast weak-supervision and interactive corrections for segmenting Whole Slide Images}%
\name{
\begin{tabular}{c}
	Antoine Habis$^{1,2}$ \qquad 
        Roy Rosman Nathanson$^{1}$ \qquad 
	Vannary Meas-Yedid$^{1}$ \qquad 
        Elsa D. Angelini$^{2}$ \qquad \\
	Jean-Christophe Olivo-Marin$^{1}$ \qquad 
\end{tabular}
}
\address{$^1$ BioImage Analysis Unit, Institut Pasteur, CNRS UMR 3691, Paris, France \\$^2$ LTCI, Télécom Paris, Institut Polytechnique de Paris, France\\}

\begin{document}
\maketitle
\begin{abstract}
This paper proposes a dynamic interactive and weakly supervised segmentation method with minimal user interactions to address two major challenges in the segmentation of whole slide histopathology images. First, the lack of hand-annotated datasets to train algorithms. Second, the lack of interactive paradigms to enable a dialogue between the pathologist and the machine, which can be a major obstacle for use in clinical routine.

We therefore propose a fast and user oriented method to bridge this gap by giving the pathologist control over the final result while limiting the number of interactions needed to achieve a good result (over $90\%$ on all our metrics with only 4 correction scribbles).

\end{abstract}
\section{Introduction}
\label{sec:intro}

Visual screening of pathological regions on whole slide images (WSIs) is a laborious and challenging task in histopathology. 
Indeed, pathological areas can be a very small part of WSI and therefore easily missed. 

Multiple instance learning (MIL), a sub-branch of weakly supervised learning,  has largely contributed to recent progress in automated WSI annotation. Recent methods \cite{mil,mil2,clam,swin} have reported very convincing results to generate heatmaps of pathological signs over large fields of views of tissues from datasets only annotated at slide-level. 

However, these methods remain black boxes which can be an obstacle to their use by clinical experts. There is a need for building more confidence with more explanations on the technology but also more control over the generated results when errors are observed.

"Human in the loop" paradigms are emerging to allow experts to contribute to the final results via a dialogue between the human and the machine. Dedicated methods are introduced so that the two help each other to improve results' quality quickly with limited user actions.

\section{related work}
In image processing, interactive segmentation tools have been used for a long time. For instance, GrabCut \cite{grabcut} enabled users to segment objects accurately with the help of bounding boxes, and intelligent scissors \cite{intelligent_scissors} made it possible for anyone to segment complex objects very precisely with just a few clicks.

More specifically in histopathology, a number of software applications have emerged, such as Tissue wand \cite{tissue_wand}, which offers an interactive tool designed for efficient annotation of gigapixel-sized histopathological images, not being constrained to a predefined annotation task, or Quick annotator \cite{qa}, which uses convolutional networks to segment many different structures using active learning with user annotations. Some methods such as DeepScribble \cite{deep_scribble} have looked at interactive segmentation for histopathology images that rely on users' scribbles. However, Tissue wand and DeepScribble both only offer local corrections and the latter is also quite costly as it uses two UNet \cite{unet} models: a first one that returns an initial segmentation and a second that performs the correction. On the other hand, Quick annotator \cite{qa} proposes to re-train an entire UNet \cite{unet} for each additional correction made by the user which can be also computationally expensive and slow.

In this work, we focus on the problem of patch-level WSI segmentation of cancer tissue images into two classes: tumor and non-tumor tissues.

We propose two original contributions: (1) training a patch-level labeling tool trained on scribbles only, (2) implementing a fast, interactive and non local correction tool using additional scribbles. 

The correction tool is based on a SVM classifier that updates the whole heatmap of the entire WSI by re-using the latent features of the initial model, making each update very fast ($\le 1s$) and cost-effective.

Our code is available online for reproducibility:
\url{https://github.com/antoinehabis/WSS-UIC}.

\section{CNN training on a dataset of scribbles}
\label{training_vgg}
\begin{figure}[h!]%
    \centering
    \subfloat[\centering Scribble of a metastatic region.]{{\includegraphics[width=2.7cm]{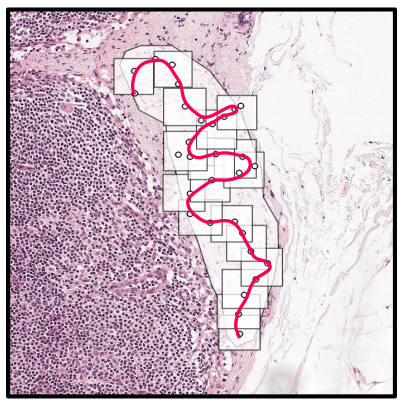}}
    \label{meta}}%
    \subfloat[\centering Scribble of a non-tumoral region]{{\includegraphics[width=5.5cm]{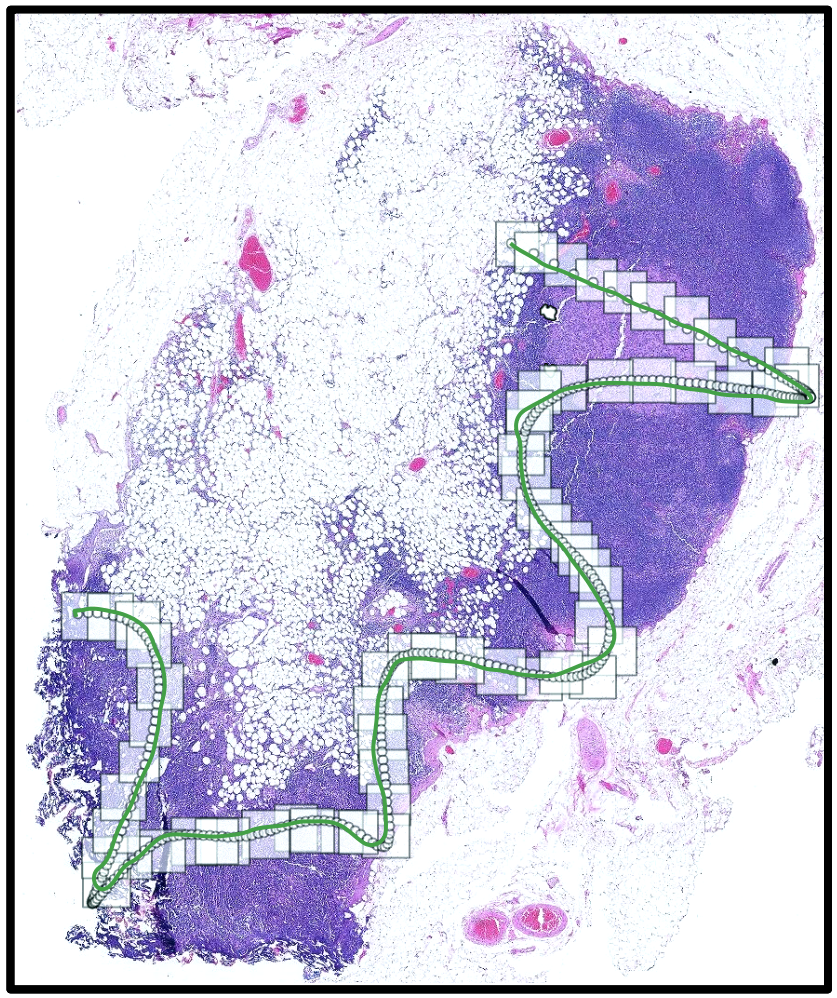} }}%
    \caption{Patch extraction along scribbles created for the 2 classes: tumoral and non-tumoral tissues.}%
    \label{scribbles}%
\end{figure}

 We train a patch-level classification network, relying on weakly supervised annotations of regions via scribbles (see \autoref{scribbles}). The user annotates individual WSIs by quickly scribbling some tumor areas and only one non-tumor area. These scribbles constitute the scribble training dataset,
 on which we train a classification network. This network provides an initial rough segmentation on the whole WSI that the user can then quickly correct with incremental additional scribbles.

\subsection{Data and pre-processing}

As an example use case we work with the H\&E Camelyon16 \cite{camelyon} dataset which contains 400  WSIs from lymph nodes annotated for 2 classes: tumor or non-tumor.  We only used n=158 WSIs containing metastases in the training and test sets, as we focus on refining the segmentation of WSIs pre-screened as cancerous. 
 
We split the sub-dataset into n=110 WSIs training, n=24 validation, and n=24 test WSIs.

\subsection{Creation of ground-truth scribbles}
\label{scribble}

Camelyon16 \cite{camelyon} dataset contains pixel-level annotations of tumour regions for each WSI. This type of annotation is very time-consuming and labour-intensive. In this first part we want to investigate whether this highly accurate annotations are necessary to train a segmentation network on WSIs. We therefore propose to create a weakly supervised scribble dataset from a fully supervised dataset of very precise per-pixel annotations. The method used to automatically generate scribbles from masks simulates the human behaviour of drawing a continuous curve along an object of interest. It is also a random process, allowing a multitude of different scribbles to be generated for a single mask. Finally, it does not exclude small overflows outside the mask and therefore takes into account potential human errors.

\begin{figure}[h!]
  \centerline{\includegraphics[width=8.9cm]{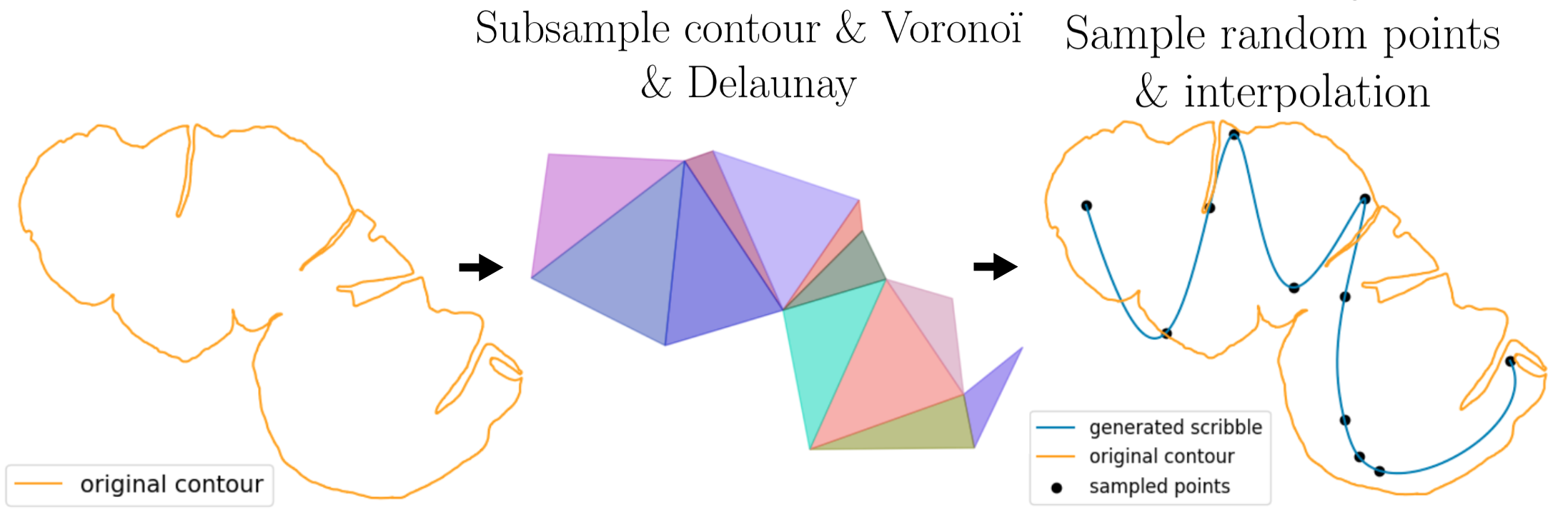}}
  \caption{Scribble creation process.}
 \label{creation_scribbles}
 \end{figure}

\indent For \textbf{tumoral scribbles}: We propose a method to artificially create random and realistic scribbles from the ground-truth masks. We sample the contours of each tumoral area into $n_c = 15$ nodes equally spaced on the contour.

We compute the Voronoï Diagram followed by the corresponding Delaunay triangulation of the inside shape and extract the longest triangle path within the shape. We randomly sample points in each triangle belonging to the longest path and interpolate to generate the scribble (See \autoref{creation_scribbles}). 

For each WSI, we generate scribbles on the top $10\%$ largest tumoral areas with a minimum of 1 and a maximum of 10. On average the number of ground-truth tumoral scribbles generated per WSI is 6.7.

For \textbf{non-tumoral scribbles}: we extract the tissue region from the background using a simple Otsu thresholding followed by a closing operation to fill the holes. We then subtract all the annotated tumoral regions from the tissue mask. We replicate the Delaunay triangulation process described above on the largest connected tissue components per WSI.

For all the WSI of our scribble dataset, we extracted a total of 203776 overlapping patches of size $W\times W = 512 \times 512$ pixels at magnification scale $\times 40$ along the generated scribbles (See \autoref{scribbles}). 

For a scribble generated with length $l$, the number of extracted patches $n_{patches}$ with spatial size $W \times W$ along the scribble and an overlap $o_v$ is then :

\begin{equation}
n_{patches} = \frac{l}{W(1 - o_v)}
\end{equation}

This corresponds to an average of 868 healthy patches and 434 tumor patches per WSI. We assigned the corresponding label (1 for tumor and 0 for healthy) to each patch to constitute our baseline training, validation, and test scribble dataset.


\begin{figure*}[ht]
  \begin{center}
  \includegraphics[width=17cm, height = 5.0cm]{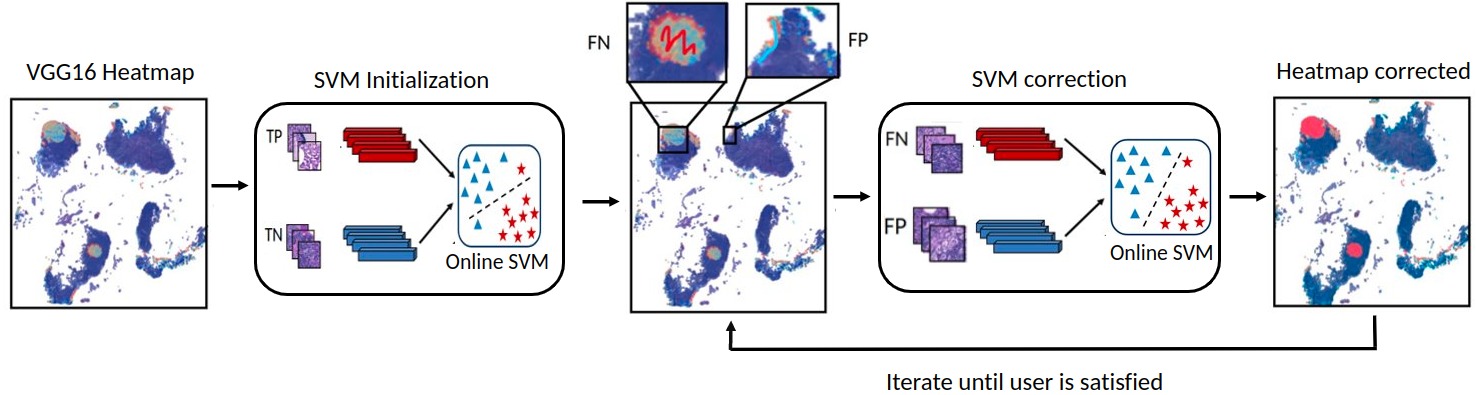}
  \caption{Incremental segmentation correction process.}
 \label{incremental_segmentation}
 \end{center}
\end{figure*}

 \subsection{Classifier: Rough initial segmentation}
\label{sec:pagestyle}

We used VGG16 \cite{VGG16} as the backbone architecture for our binary patch classifier and used dropout regularization with a probability of $0.2$. We started with weights pre-trained on ImageNet \cite{imagenet} that we fine-tuned for 10 epochs on our train scribble dataset. We used Adam \cite{Adam} optimizer and an initial learning rate of $l_r = 10^{-3}$. For augmentations, we use flips, $90^{\circ}$ rotations, and small translations.

We optimized the threshold used on the output probability map (heatmap) to maximize the $F1$ score on the scribble validation dataset. This led to $t_{thresh} = 0.33$ for an overlap value $o_v = 50\%$.

\section{Incremental segmentation corrections}

Once the VGG16 is trained on our weakly supervised dataset, the results can be observed on the test images and any potential errors in the prediction of the weakly trained model can be corrected using additional scribbles.

\subsection{Naïve implementation}
\label{naive}

Our iterative correction process is illustrated in \autoref{incremental_segmentation}. We use a SVM classifier working on the latent features extracted from the previously trained VGG16 \cite{VGG16} to enable interactive corrections. The SVM has the advantage of having very few parameters so that it is very fast to train but also completely adapted to incremental learning.

\noindent \textbf{1. SVM Initialization:}  "true-positive (TP)" and "true-negative (TN)" patches are extracted as the patches with the highest and lowest scores respectively above and below $t_{thresh}$. Per WSI, we extract a maximum of 1,000 patches for each class to have a fast ($\le 1s$) initialization step and avoid class imbalance.

\noindent \textbf{2. SVM Correction:} 
Additional input scribbles are provided by the user over regions of the heatmap considered misclassified. False-positives (FP) and false-negatives (FN) patches are extracted from the corrective scribbles.
In our approach, we automatically generate scribbles on FP or FN areas with only $n_{patches}=10$ FP and $n_{patches}=10$ FN patches per WSI to minimize the expected size of the correction scribble.  We refit the SVM for $n_{epoch}$ and update all WSI patches with the SVM predictions except the patches along the scribbles that are hard coded. This process can be iterated $n_{pass}$ passes.

The hyperparameter $n_{epoch}$ reflects the balance to be found between the importance of corrections and the reliability of the initial segmentation. Indeed, if the initial segmentation is almost perfect, then we would like corrections to have little impact on the final result, and vice versa. We, therefore, defined its optimal value $n_{epoch}^*$ as the one returning the highest mean $F_1$ score over the validation set after $n_{pass}$ correction passes.

\subsection{Uncertainty-based implementation}
It is desirable to adjust the level of SVM retraining per WSI (controlled by a fixed value of $n_{epoch}$ in \ref{naive}) with the initial quality of the rough segmentation.  

To infer the quality of a given VGG16 \cite{VGG16} prediction without relying on a ground-truth annotation, we use the notion of uncertainty from Monte-Carlo dropout introduced in \cite{mc} that have already shown good results in \cite{mc_app1, mc_app2}. More specifically in histology, few papers have already explored its effectiveness and reliability \cite{mcua, thagaard, uncertainty}.

First, we run $n_{MC}$ predictions for each patch using Monte-Carlo dropout on VGG16 \cite{VGG16} trained weights from \autoref{training_vgg} . Our observations showed that large uncertainty values are mainly located in tumor areas. Thus, in order to have a WSI-level uncertainty measure that is independent of the size of the tumor we introduce the following uncertainties: 

\begin{equation}
\begin{split}
     H_{\text{WSI}}      & = \frac{1}{|T|} \sum_{\mathcal{P} \in T} H(X_{\mathcal{P}}) \\
     \sigma_{\text{WSI}} & = \frac{1}{|T|} \sum_{\mathcal{P} \in T} \sigma(X_{\mathcal{P}})
\end{split}
\label{Eq1}
\end{equation}

\noindent with  $H$ the entropy, $\sigma$ the standard deviation, $\mathcal{P}$ a patch in the WSI, $X_{\mathcal{P}}$ the vector of $n_{MC}$ predictions on $\mathcal{P}$ and   $T = \left\{\mathcal{P} \in \text{WSI}, E[X_{\mathcal{P}}]>t_{thresh} \right\}$

To evaluate the reliability of these WSI-level uncertainty measures to infer the overall segmentation quality, we evaluated their linear correlation with the WSI-level $F_1$ scores of the VGG16 labels on the test and validation subsets.

\begin{figure}[ht]
  \includegraphics[width=\columnwidth]{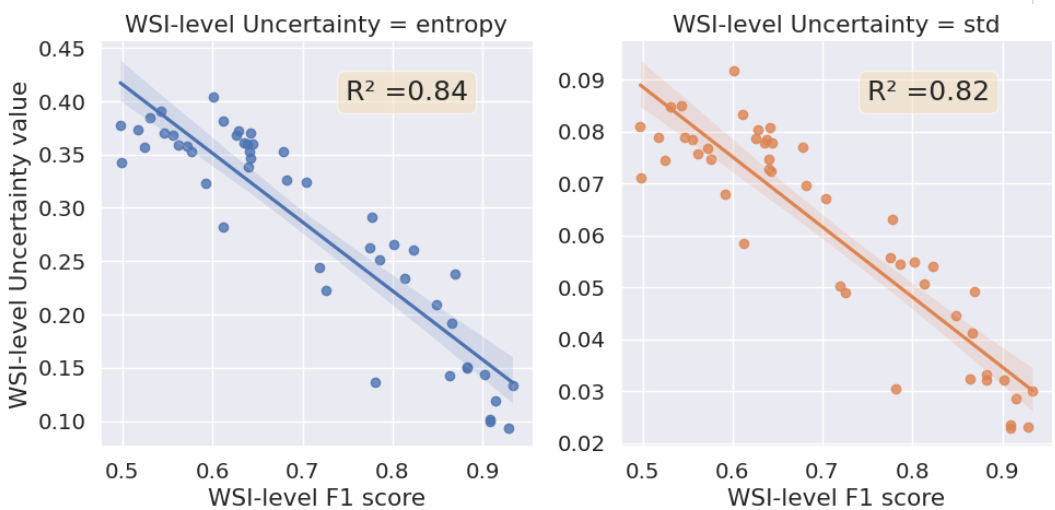}
  \caption{WSI-level uncertainty measures versus $F_1$ scores of the VGG16 compared with the ground truth calculated on all the WSI patches.}
  \label{uncertainties}

\end{figure}

\autoref{uncertainties} shows that the two uncertainty measures have a linear relation with the $F_1$ scores and correlations above 0.82. They are therefore both good unsupervised proxies of the segmentation quality, with $H_{\text{WSI}}$ having a higher correlation. We therefore propose to set a WSI-specific value of $n_{epoch}$ based on $H_{\text{WSI}}$  using a window centered on $n_{epoch}^*$: 
\begin{equation}
    n_{epoch}^{\text{WSI}} = 2 \times H^*_{\text{WSI}} \times n_{epoch}^*
\end{equation}

\noindent where $H^*_{\text{WSI}}$ is the normalised value of $H_{\text{WSI}}$ in $[0,1]$ using the max and min from \autoref{uncertainties}. Thus, the greater the uncertainty regarding VGG16 segmentation, the greater the impact of corrections. On the contrary, if the uncertainty is very low, the impact of corrections will also be low, as the user does not want to completely alter the whole heatmap with the provided corrections.

\section{Results}

For the naïve approach, we tested the following values of $n_{epoch} \in \left\{1, 5, 10, 20, 30, 40, 50, 60,70, 80, 90, 100 \right\}$.

For $n_{pass}=1$, the best F1 value on the validation set was found with $n_{epoch}^{*}$ = 20 and for $n_{pass} = 2,3,4$ with $n_{epoch}^{*}$ = 30.

In the remaining, we detail results on the test set for $n_{pass}= 4$ correction passes with $n_{epoch}^{*} = 30$.

For the uncertainty-based approach, we set $n_{MC}$ = 20.
For each test WSI (n=24) we ran the whole correction process 10 times using randomly chosen correction scribbles. This leads to 10 segmentations per WSI per correction pass. 

We report in \autoref{naive_table} for each performance metric its overall average value over the $10 \times 24$ segmentations and the average of its standard deviation per WSI across the 10 correction runs.

\begin{table}[ht]
\caption{WSI-level classification quality metrics on the test set (n=24). VGG16 = weakly-supervised rough segmentation. Other lines report scores after Naïve incremental segmentation corrections using input corrective scribbles and SVM classifiers with $n_{pass} = 4$ and $n_{epoch}^{*} = 30$.}
\label{naive_table}

\scalebox{0.85}{
\begin{tabular}{l|l|l|l|l|}
\cline{2-5}
                                     & \begin{tabular}[c]{@{}l@{}}Balanced \\ Accuracy(\%)\end{tabular} & \begin{tabular}[c]{@{}l@{}}mean \\ Precision(\%)\end{tabular} & \begin{tabular}[c]{@{}l@{}}mean\\ Recall(\%)\end{tabular} & \begin{tabular}[c]{@{}l@{}}F1\\ score(\%)\end{tabular} \\ \hline
\multicolumn{1}{|l|}{VGG16}          & $75.4$                                                             & $71.0$                                                          & $79.7$                                                      & $74.2$                                                   \\ \hline
\multicolumn{1}{|l|}{$1^{st}$ pass} & $78.2 \pm 0.1$                                                       & $70.5 \pm 0.1$                                                    & $86.0 \pm 0.2$                                                & $76.4 \pm 0.1$                                             \\ \hline
\multicolumn{1}{|l|}{$2^{nd}$ pass} & $81.7 \pm 0.1$                                                       & $74.3 \pm 0.1$                                                    & $89.0 \pm 0.2$                                                & $79.7 \pm 0.1$                                             \\ \hline
\multicolumn{1}{|l|}{$3^{rd}$ pass} & $84.0 \pm 0.3$                                                       & $79.2 \pm 0.7$                                                    & $88.8 \pm 0.2$                                                & $82.4 \pm 0.4$                                             \\ \hline
\multicolumn{1}{|l|}{$4^{th}$ pass} & \textbf{88.9}$\pm$ 0.4                                                        & \textbf{87.5} $\pm$ 0.7                                                    & \textbf{90.3} $\pm$ 0.2                                                & \textbf{87.7} $\pm$ 0.4                                             \\ \hline
\end{tabular}}
\end{table}

\begin{table}[ht]
\caption{Same experiment as in \autoref{naive_table} but using uncertainty-based incremental segmentation correction}
\label{mc_table}
\scalebox{0.85}{
\begin{tabular}{l|l|l|l|l|}
\cline{2-5}
                                     & \begin{tabular}[c]{@{}l@{}}Balanced \\ Accuracy(\%)\end{tabular} & \begin{tabular}[c]{@{}l@{}}mean \\ Precision(\%)\end{tabular} & \begin{tabular}[c]{@{}l@{}}mean\\ Recall(\%)\end{tabular} & \begin{tabular}[c]{@{}l@{}}F1\\ score(\%)\end{tabular} \\ \hline
\multicolumn{1}{|l|}{VGG16}          & $75.4$                                                             & 71.0                                                          & 79.7                                                      & $74.2$                                                   \\ \hline
\multicolumn{1}{|l|}{$1^{st}$ pass} & 77.9 $\pm$ 0.1                                                       & 69.3 $\pm$ 0.1                                                    & 86.4 $\pm$ 0.2                                                & 75.7 $\pm$ 0.1                                            \\ \hline
\multicolumn{1}{|l|}{$2^{nd}$ pass} & 81.6 $\pm$ 0.2                                                       & 74.7 $\pm$ 0.3                                                    & 88.4 $\pm$ 0.2                                               & 79.7 $\pm$ 0.3                                             \\ \hline
\multicolumn{1}{|l|}{$3^{rd}$ pass} & 86.6 $\pm$ 0.4                                                       & 85.3 $\pm$ 0.6                                                    & 87.9 $\pm$ 0.4                                                & 85.4 $\pm$ 0.5                                             \\ \hline
\multicolumn{1}{|l|}{$4^{th}$ pass} & \textbf{90.9} $\pm$ 0.7                                                       & \textbf{91.4} $\pm$ 1.4                                                    & \textbf{90.5} $\pm$ 0.1                                                & \textbf{90.0} $\pm$ 0.8                                             \\ \hline
\end{tabular}}
\end{table}

\begin{figure}[ht]
  \includegraphics[width=\columnwidth]{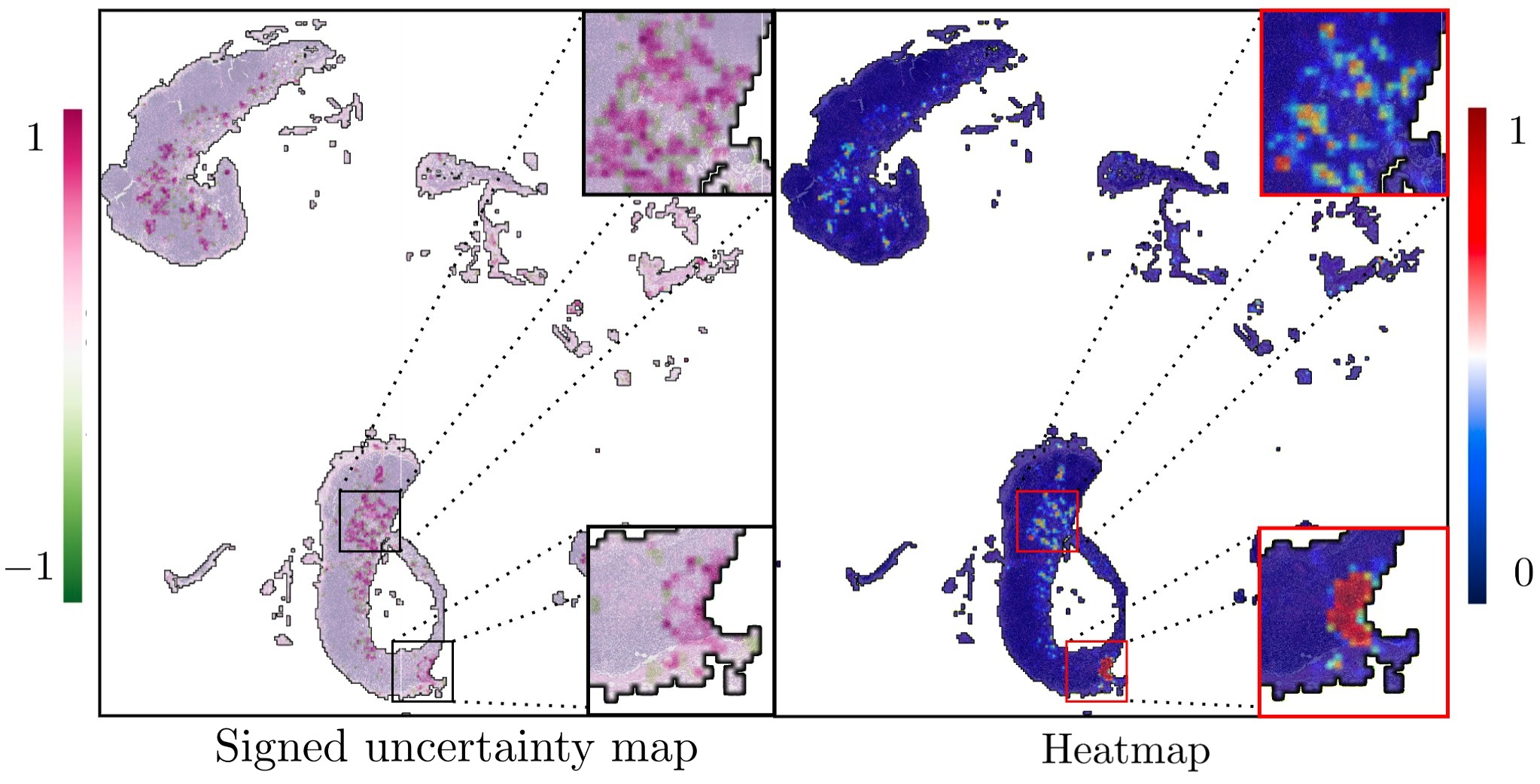}
    \caption{Visualization of the signed uncertainty map with $t_{thresh} = 0.33$ (left) and heatmap (right) of one WSI. Zoomed regions show areas with TPs (bottom right)and FPs (top right). The signed uncertainty map shows high uncertainty values in the FP region and rather low values in the TP one.}
  \label{heatmaps}
\end{figure}

Results using the uncertainty-based corrections are reported in \autoref{mc_table}. 
We observe better performance with the uncertainty-based corrections than with the naïve setup after the 3rd and 4th passes. Indeed, all classification metrics are above $90\%$ after the 4th corrections and we therefore observe a $17\%$ improvement in the efficiency of the corrector between the uncertainty-based implementation and the naïve implementation. This performance compares well with DeepScribble \cite{deep_scribble} while tested for another type of tumor. Indeed they reached optimal performance (IoU $\approx 0.9$ in their case) with an average of 5.9 corrective scribbles.

Another major advantage of computing patch-level uncertainties is the generation of a signed uncertainty heatmap. The signed uncertainty $H(X_{\mathcal{P}})^{sign}$ of a patch $\mathcal{P}$, is defined as:

\begin{equation}
      H(X_{\mathcal{P}})^{sign} = \sign (E[X_{P}] - t_{thresh}) \times H(X_{\mathcal{P}})
\end{equation}
\noindent where $E[]$ is the mean operator.
The signed uncertainty map as shown in \autoref{heatmaps} (left part) highlights risky areas with high uncertainty (i.e. low confidence in the rough segmentation) which can guide the user when selecting areas to scribble.
Values close to -1 are potential FN and values close to 1 are potential FP.

\section{Conclusion}
\label{conclusion}

In this paper, we proposed a new paradigm for interactive scribble-based corrections of a rough binary segmentation of tumoral regions on WSI slides.
Our results on the Camelyon16 \cite{camelyon} dataset show that few additional annotations enable to reach scores up to 90 \% on all our metrics. 
Our proposed correction procedure is based on a SVM working in the latent space of the rough classifier. It is therefore very fast ($\approx 1s$) as it does not require recalculating latent features at each step contrary to \cite{deep_scribble}.

In particular, we show that Monte Carlo dropout on a weakly-supervised VGG16 \cite{VGG16} can generate patch-level uncertainty measures that have 3 usages: (1) infer the WSI-level quality of the provided rough segmentation having high correlation with F1 scores, (2) improve the proposed correction process by tuning the importance of the correction with regard to the initial segmentation, (3) guide the user to position the corrective scribbles via the signed uncertainty map.

\section{Acknowledgments}
\label{sec:print}

This project has received funding from the Innovative Medicines Initiative 2 Joint Undertaking under grant agreement No 945358. This Joint Undertaking receives support from the European Union’s Horizon 2020 research and innovation program and EFPIA.www.imi.europe.eu

\bibliographystyle{IEEEbib}
\bibliography{refs}

\begin{thebibliography}{10}

\bibitem{mil}
Yan Xu, Jun-Yan Zhu, Eric Chang, Maode Lai, and Z.~Tu,
\newblock ``Weakly supervised histopathology cancer image segmentation and classification,''
\newblock {\em Medical Image Analysis}, vol. 18, pp. 591--604, 02 2014.

\bibitem{mil2}
Xi~Wang, Hao Chen, Caixia Gan, Huangjing Lin, Qi~Dou, Efstratios Tsougenis, Qitao Huang, Muyan Cai, and Pheng-Ann Heng,
\newblock ``Weakly supervised deep learning for whole slide lung cancer image analysis,''
\newblock {\em IEEE Transactions on Cybernetics}, vol. 50, pp. 3950--3962, 2020.

\bibitem{clam}
Ming~Y Lu, Drew~FK Williamson, Tiffany~Y Chen, Richard~J Chen, Matteo Barbieri, and Faisal Mahmood,
\newblock ``Data-efficient and weakly supervised computational pathology on whole-slide images,''
\newblock {\em Nature Biomedical Engineering}, vol. 5, no. 6, pp. 555--570, 2021.

\bibitem{swin}
Ziniu Qian, Kailu Li, Maode Lai, Eric I-Chao Chang, Bingzheng Wei, Yubo Fan, and Yan Xu,
\newblock ``Transformer based multiple instance learning for weakly supervised histopathology image segmentation,''
\newblock in {\em Medical Image Computing and Computer Assisted Intervention (MICCAI)}, 2022, pp. 160--170.

\bibitem{grabcut}
Carsten Rother, Vladimir Kolmogorov, and Andrew Blake,
\newblock ``{GrabCut}: {I}nteractive foreground extraction using iterated graph cuts,''
\newblock {\em ACM Trans. Graph.}, vol. 23, no. 3, pp. 309--314, 2004.

\bibitem{intelligent_scissors}
Eric~N. Mortensen and William~A. Barrett,
\newblock ``Interactive segmentation with intelligent scissors,''
\newblock {\em Graphical Models and Image Processing}, vol. 60, no. 5, pp. 349--384, 1998.

\bibitem{tissue_wand}
Martin LindvaN, Alexander Sanner, Fredrik Petre, Karin Lindman, Darren Treanor, Claes Lundstrbm, and Jonas Ldwgren,
\newblock ``Tissue{W}and, a rapid histopathology annotation tool,''
\newblock {\em Journal of Pathology Informatics}, vol. 11, no. 1, pp. 27, 2020.

\bibitem{qa}
Runtian Miao, Robert Toth, Yu~Zhou, Anant Madabhushi, and Andrew Janowczyk,
\newblock ``{Quick Annotator: an open‐source digital pathology based rapid image annotation tool},''
\newblock {\em The Journal of Pathology: Clinical Research}, vol. 7, no. 6, pp. 542--547, 2021.

\bibitem{deep_scribble}
Sungduk Cho, Hyungjoon Jang, Jing wei Tan, and Won-Ki Jeong,
\newblock ``Deepscribble: Interactive pathology image segmentation using deep neural networks with scribbles,''
\newblock in {\em IEEE International Symposium on Biomedical Imaging (ISBI)}, 2021, pp. 761--765.

\bibitem{unet}
Weihao Weng and Xin Zhu,
\newblock ``{UNet: Convolutional Networks for Biomedical Image Segmentation},''
\newblock {\em IEEE Access}, vol. 9, pp. 16591--16603, 2021.

\bibitem{camelyon}
Geert Litjens, Peter Bandi, Babak Ehteshami Bejnordi, Oscar Geessink, Maschenka Balkenhol, Peter Bult, Altuna Halilovic, Meyke Hermsen, Rob van de Loo, Rob Vogels, Quirine~F Manson, Nikolas Stathonikos, Alexi Baidoshvili, Paul van Diest, Carla Wauters, Marcory van Dijk, and Jeroen van der Laak,
\newblock ``{1399 H\&E-stained sentinel lymph node sections of breast cancer patients: the CAMELYON dataset},''
\newblock {\em GigaScience}, vol. 7, no. 6, pp. giy065, 05 2018.

\bibitem{VGG16}
Karen Simonyan and Andrew Zisserman,
\newblock ``Very deep convolutional networks for large-scale image recognition,''
\newblock {\em arXiv preprint arXiv:1409.1556}, 2014.

\bibitem{imagenet}
Jia Deng, Wei Dong, Richard Socher, Li-Jia Li, Kai Li, and Li~Fei-Fei,
\newblock ``Imagenet: A large-scale hierarchical image database,''
\newblock in {\em IEEE Conference on Computer Vision and Pattern Recognition (CVPR)}, 2009, pp. 248--255.

\bibitem{Adam}
Diederik~P Kingma and Jimmy Ba,
\newblock ``Adam: A method for stochastic optimization,''
\newblock {\em arXiv preprint arXiv:1412.6980}, 2014.

\bibitem{mc}
Yarin Gal and Zoubin Ghahramani,
\newblock ``Dropout as a {B}ayesian approximation: {R}epresenting model uncertainty in deep learning,''
\newblock in {\em International Conference on Machine Learning (ICML) - Volume 48}, 2016, p. 1050–1059.

\bibitem{mc_app1}
Lavsen Dahal, Aayush Kafle, and B.~Khanal,
\newblock ``Uncertainty estimation in deep {2D} echocardiography segmentation,''
\newblock {\em ArXiv}, vol. abs/2005.09349, 2020.

\bibitem{mc_app2}
Tanya Nair, Doina Precup, Douglas~L. Arnold, and Tal Arbel,
\newblock ``Exploring uncertainty measures in deep networks for multiple sclerosis lesion detection and segmentation,''
\newblock {\em Medical Image Analysis}, vol. 59, pp. 101557, 2020.

\bibitem{mcua}
Zakaria Senousy, Mohammed~M Abdelsamea, Mohamed~Medhat Gaber, Moloud Abdar, U~Rajendra Acharya, Abbas Khosravi, and Saeid Nahavandi,
\newblock ``Mcua: Multi-level context and uncertainty aware dynamic deep ensemble for breast cancer histology image classification,''
\newblock {\em IEEE Transactions on Biomedical Engineering}, vol. 69, no. 2, pp. 818--829, 2021.

\bibitem{thagaard}
Jeppe Thagaard, S{\o}ren Hauberg, Bert van~der Vegt, Thomas Ebstrup, Johan~D Hansen, and Anders~B Dahl,
\newblock ``Can you trust predictive uncertainty under real dataset shifts in digital pathology?,''
\newblock in {\em Medical Image Computing and Computer Assisted Intervention (MICCAI)}, 2020, pp. 824--833.

\bibitem{uncertainty}
Sidi Yang and Thomas Fevens,
\newblock ``Uncertainty quantification and estimation in medical image classification,''
\newblock in {\em International Conference on Artificial Neural Networks (ICANN)}, 2021, pp. 671--683.

\end{thebibliography}

\end{document}